\documentclass[10pt,twocolumn,letterpaper]{article}

\usepackage{cvpr}              % To produce the CAMERA-READY version
\usepackage{amsmath}
\usepackage{amssymb}
\usepackage{booktabs}
\usepackage{multicol}
\usepackage{multirow}
\usepackage{textcomp}
\usepackage{stfloats}
\newsavebox{\measurebox}
\usepackage{graphicx}

\usepackage{bm}
\usepackage{caption}
\usepackage{subcaption}
\usepackage[pagebackref,breaklinks,colorlinks]{hyperref}

\usepackage[capitalize]{cleveref}
\crefname{section}{Sec.}{Secs.}
\Crefname{section}{Section}{Sections}
\Crefname{table}{Table}{Tables}
\crefname{table}{Tab.}{Tabs.}

\def\cvprPaperID{5153} % *** Enter the CVPR Paper ID here
\def\confName{CVPR}
\def\confYear{2023}

\begin{document}

\title{FAQ: Feature Aggregated Queries for Transformer-based\\ Video Object Detectors}

\author{Yiming Cui\\
University of Florida\\
{\tt\small cuiyiming@ufl.edu}
\and
Linjie Yang\\
ByteDance Inc.\\
{\tt\small yljatthu@gmail.com}
% \and 
% Ding Liu \\
% ByteDance Inc.\\
% {\tt\small liudingdavy@gmail.com}
}
\maketitle

%%%%%%%%% ABSTRACT
\begin{abstract}
Video object detection needs to solve feature degradation situations that rarely happen in the image domain. One solution is to use the temporal information and fuse the features from the neighboring frames. With Transformer-based object detectors getting a better performance on the image domain tasks, recent works began to extend those methods to video object detection. However, those existing Transformer-based video object detectors still follow the same pipeline as those used for classical object detectors, like enhancing the object feature representations by aggregation. In this work, we take a different perspective on video object detection. In detail, we improve the qualities of queries for the Transformer-based models by aggregation. To achieve this goal, we first propose a vanilla query aggregation module that weighted averages the queries according to the features of the neighboring frames. Then, we extend the vanilla module to a more practical version, which generates and aggregates queries according to the features of the input frames. Extensive experimental results validate the effectiveness of our proposed methods: On the challenging ImageNet VID benchmark, when integrated with our proposed modules, the current state-of-the-art Transformer-based object detectors can be improved by more than $2.4\%$ on mAP and $4.2\%$ on AP$_\text{50}$.
\end{abstract}

%%%%%%%%% BODY TEXT
\section{Introduction}
\begin{figure*}[!bt]
    \centering
    % \begin{subfigure}[b]{0.45\textwidth}
    %      \centering
    %      \includegraphics[angle=270,origin=c, width=6cm, ]{img/detr2.png}
    %      \caption{Transformer-based object detectors for the image domain}
    %  \end{subfigure}
    %       \begin{subfigure}[b]{0.45\textwidth}
    %      \centering
    %      \includegraphics[width=7cm]{latex/img/featureAgg2.png}
    %      \caption{Post-processing based video object detectors}
    %  \end{subfigure}
    %  \begin{subfigure}[b]{0.45\textwidth}
    %      \centering
    %      \includegraphics[width=7cm]{latex/img/postprocess2.png}
    %      \caption{Feature-aggregation based video object detectors}
    %  \end{subfigure}
    %  \begin{subfigure}[b]{0.45\textwidth}
    %      \centering
    %      \includegraphics[width=6cm]{latex/img/ours2.png}
    %      \caption{Our proposed method}
    %  \end{subfigure}
     \includegraphics[width=16cm]{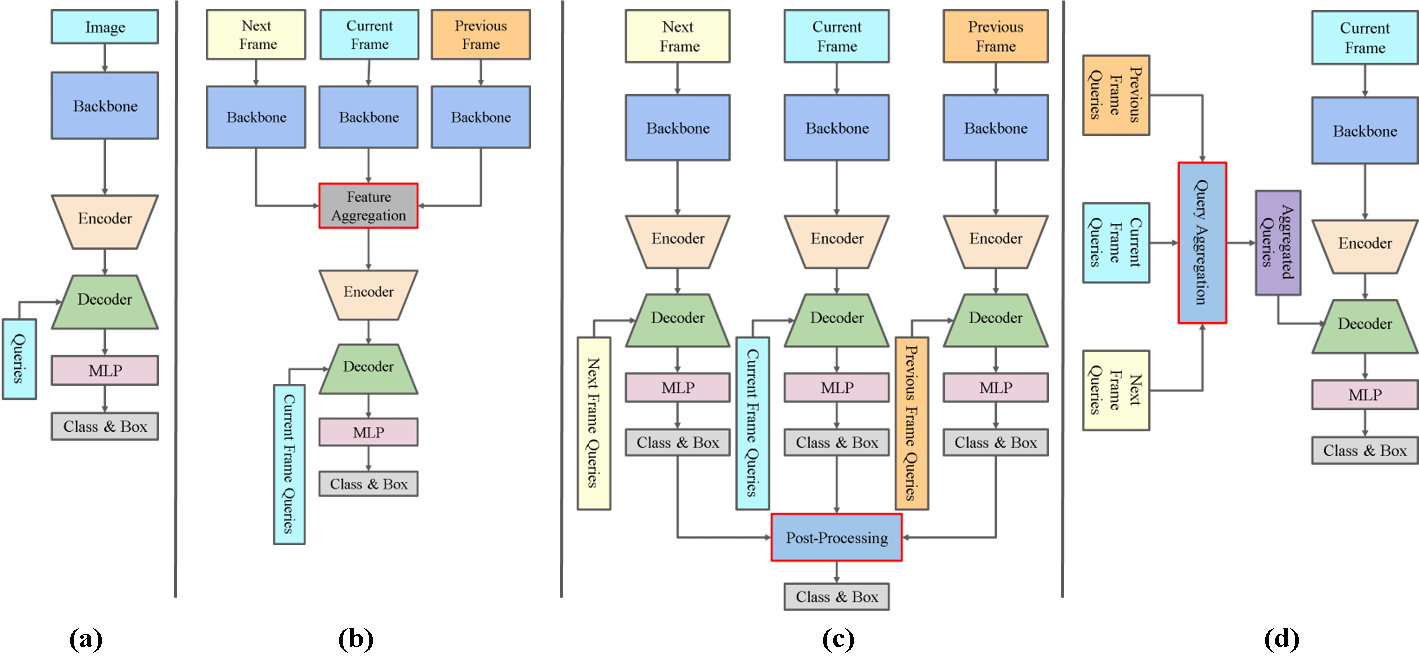}
    \caption{The differences between the existing works and ours. (a) Transformer-based object detectors. (b) Feature-aggregation based video object detectors. (c) Post-processing based video object detectors. (d) Ours. Previous works can be divided into feature-aggregation based (b) and post-processing based (c) models. For Transformer-based models, these works either enhance the features used for detection or the prediction results of each frame. In contrast, our methods (d) pay attention to the aggregation of queries for those Transformer-based object detection models to handle the feature degradation issues.}
    \label{fig: methodCompare}
\end{figure*}

Object detection is an essential yet challenging task which aims to localize and categorize all the objects of interest in a given image. With the development of deep learning, extraordinary processes have been achieved in static image object detection \cite{Cai_2019, sun2021sparse, duan2019centernet, law2018cornernet, lin2017focal}. Existing object detectors can be mainly divided into three categories: two-stage \cite{ren2015faster, he2017mask, Cai_2019, girshick2015fast, lin2017feature}, one-stage \cite{tian2019fcos, redmon2016you, redmon2017yolo9000, redmon2018yolov3, lin2017focal, liu2021visual, tian2021fcos, liu2016ssd} and query-based models \cite{detr, sun2021sparse, roh2021sparse, zhu2021deformable,liu2022dab,gao2022adamixer}. For better performance, two-stage models generate a set of proposals and then refine the prediction results, like R-CNN families \cite{gao2021fast, dai2016r, ren2015faster, he2017mask}. However, these two-stage object detectors usually suffer from a low inference speed. Therefore, one-stage object detectors are introduced to balance the efficiency and performance, which directly predicts the object locations and categories based on the input image feature maps, like YOLO series \cite{redmon2016you, shafiee2017fast, redmon2018yolov3, redmon2017yolo9000} and FCOS \cite{tian2019fcos, tian2021fcos}. Recently, query-based object detectors have been introduced, which generate the predictions based on a series of input queries and do not require complicated post-processing pipelines like NMS \cite{liu2019adaptive, neubeck2006efficient, bodla2017soft}. Some typical example models are DETR series \cite{detr, zhu2021deformable, liu2022dab, roh2021sparse} in Figure \ref{fig: methodCompare}(a) and Sparse R-CNN series \cite{sun2021sparse, Fang_2021_ICCV, hong2022dynamic}.

With the existing approaches getting better performance on the image domain, researchers began to extend the tasks to the video domain \cite{yang2019video, russakovsky2015imagenet, kim2020video,xu2018youtube, voigtlaender2019mots}. One of the most challenging issues of video object detection is handling the feature degradation caused by motion, which rarely appears in static images. Since videos provide informative temporal hints, post-processing-based video object detectors are proposed \cite{han2016seq, Kang_2016,sabater2020robust, kang2017t, belhassen2019improving}. As shown in Figure \ref{fig: methodCompare}(c), these methods first apply image object detectors on every individual frame and then associate the prediction results. However, since the image object detectors and the post-processing pipelines are not optimized jointly, these models usually suffer from poor performance.

Besides post-processing methods, feature-aggregation-based models \cite{wu2019sequence, cui2021tf, zhu2017flow, chen2020memory, zhou2022transvod, he2021end, jiang2020learning, gong2021temporal} are introduced to improve the feature representations for video object detection. These approaches first weighted average the features from the neighboring frames and then fed the aggregated features into the task heads for the final prediction, as shown in Figure \ref{fig: methodCompare}(b). The pipeline for weighted averaging is usually based on feature similarity \cite{zhu2017flow, wu2019sequence, zhu2017deep, chen2020memory, wang2018fully} or learnable networks \cite{cui2021tf, he2021end, zhou2022transvod}. Since Transformer-based models perform better on image object detection, researchers have begun extending them to the video domain \cite{zhou2022transvod, he2021end,wang2022ptseformer}. TransVOD families \cite{zhou2022transvod, he2021end} introduce a temporal Transformer to the original Deformable-DETR \cite{zhu2021deformable} to fuse both the spatial and temporal information to handle the feature degradation issue. Similarly, PTSEFormer \cite{wang2022ptseformer} introduces progressive feature aggregation modules to the current Transformer-based image object detectors to boost the performance. Following the TransVOD series \cite{zhou2022transvod, he2021end}, we use Transformer-based object detectors as the baseline models in this work.

Unlike the existing models, we take a deeper look at the Transformer-based object detectors and find out the unique properties of their designs. We notice that the queries of Transformer-based object detectors play an essential role in the final prediction performance. Therefore, different from the existing works, which apply different modules to aggregate features (Figure \ref{fig: methodCompare}(b)) or detection results in every single frame (Figure \ref{fig: methodCompare}(c)), we introduce a module to aggregate the queries for the Transformer decoder, as shown in Figure \ref{fig: methodCompare}(d). The existing TransVOD families \cite{zhou2022transvod, he2021end} initialize the spatial and temporal queries randomly regardless of the input frames and then aggregate them after several Transformer layers. Unlike them, our models focus on initializing the object queries and enhancing their qualities of Transformer-based approaches for better performance. By associating and aggregating the initialization of the queries with the input frames, our models can achieve a much better performance compared to the TransVOD families \cite{zhou2022transvod, he2021end} and PTSEFormer \cite{wang2022ptseformer}. Meanwhile, our methods can be integrated into most of the existing Transformer-based image object detectors to be adaptive to the video domain task.  Our contributions are summarized as follows:
\begin{itemize}
    \item To the best of our knowledge, we are the first to focus on the initialization of queries and aggregate them based on the input features for Transformer-based video object detectors to balance the model efficiency and performance.
    \item We design a vanilla query aggregation (VQA) module, which enhances the query representations for the Transformer-based object detectors to improve their performance on the video domain tasks. Then we extend it to a dynamic version, which can adaptively generate the initialization of queries and adjust the weights for query aggregation according to the input frames.
    \item Our proposed method is a plug-and-play module which can be integrated into most of the recent state-of-the-art Transformer-based object detectors for video tasks. Evaluated on the ImageNet VID benchmark, the performance of video object detection can be improved by at least $2.0\%$ on mAP when integrated with our proposed modules.
\end{itemize}
\section{Related Works}
\noindent\textbf{Image object detectors.} Image object detection requires the model to accurately predict the location and category of the objects in the input image \cite{liu2016ssd, duan2019centernet,law2018cornernet, zhang2021studying, zhang2022attention}. R-CNN families \cite{dai2016r, he2017mask, ren2015faster, ren2015faster} introduce the basic framework for two-stage object detectors, where proposals are first roughly predicted and then refined for better performance. Following R-CNN families, multiple two-stage works \cite{Cai_2019, damen2018scaling, lin2017feature, dai2016r} are proposed to improve the performance of two-stage object detectors. Besides the detection accuracy, one-stage object detectors \cite{liu2016ssd, tian2019fcos, lin2017focal, tian2021fcos, liu2021sg} are introduced, which generate the predictions without the need for region proposals to improve the inference speeds. Meanwhile, anchor-free based models \cite{tian2019fcos, tian2021fcos, duan2020corner, piao2022accloc} and point-based methods \cite{law2018cornernet, duan2019centernet,zhou2019bottom} are proposed, which provides different perspectives to the object detection fields. However, all the models mentioned above require the pre-processing pipeline like anchor design \cite{ren2015faster, liu2016ssd, Cai_2019} or post-processing like NMS \cite{liu2019adaptive, neubeck2006efficient}.

Recently, query-based object detectors \cite{gao2022adamixer, roh2021sparse, sun2021sparse, detr, zhu2021deformable, liu2022dab, Fang_2021_ICCV, hu2021istr, dong2021solq, dai2021dynamic, wang2022anchor, yao2021efficient} are proposed, which removes the complicated anchor designs or post-processing pipelines in the traditional models. Among them, Transformer-based models, especially DETR \cite{detr} is a pioneer work which treats the object detection task as a bipartite matching problem and introduces a sequence-to-sequence model with $100$ randomly initialized queries. Following DETR, multiple works \cite{zhu2021deformable, liu2022dab, roh2021sparse, Meng_2021_ICCV, gao2021fast, zhang2022dino, chen2022group, li2022dn, chen2022group, jia2022detrs} are proposed to improve the inference speed, performance, or convergence efficiency of DETR. For one direction \cite{gao2021fast, Meng_2021_ICCV, liu2022dab, yao2021efficient, wang2022anchor}, prior knowledge is introduced to the queries for faster convergence. In another way, modulated self-attention modules with fewer operations are proposed to improve the convergence speed of DETR \cite{roh2021sparse, zhu2021deformable, liu2022dab, dai2021dynamic, zhang2021detr}. Meanwhile, multiple training strategies \cite{zhang2022dino, li2022dn, chen2022group, jia2022detrs} are introduced to improve the convergence speed and performance of the DETR series models. In this work, we mainly focus on Transformer-based image object detectors and design modules to extend them for the tasks in the video domain.

\noindent\textbf{Video object detectors.} Depending on the pipeline to improve the detection performance, existing methods can be divided into post-processing \cite{han2016seq, sabater2020robust, Kang_2016, kang2017t} and feature-aggregation \cite{wu2019sequence, cui2021tf, chen2020memory, zhu2017flow, wang2018fully, zhou2022transvod, he2021end, wang2022ptseformer} based models. Post-processing based video object detectors extend the models from the image domain by merging the prediction results according to the temporal information. For example, Seq-NMS \cite{han2016seq} associates the bounding boxes from different frames with the IoU threshold; T-CNN \cite{Kang_2016} links the object detection results from each frame according to optical flows. These methods are not trained end-to-end, and the performances are always sub-optimal. Feature-aggregation based models enhance the feature representations of the current frame by fusing those from the neighboring frames. FGFA \cite{zhu2017flow} and MANet \cite{wang2018fully} weight average the features from the neighboring frames after being warped based on optical flows. SELSA \cite{wu2019sequence}, and Temporal ROI Align \cite{gong2021temporal} fuse the neighboring frames according to their semantic similarity. MEGA \cite{chen2020memory} takes both temporal information and semantic similarity into account and aggregates local and global features jointly. Different from the methods mentioned above, which generate weights for aggregation using cosine similarity, TF-Blender \cite{cui2021tf} applies a learnable network to predict the weights. 

All the methods mentioned above are designed for the classic object detectors like Faster R-CNN \cite{ren2015faster}, or CenterNet \cite{duan2019centernet}. Recently, with DETR series \cite{detr, gao2021fast, zhu2021deformable, roh2021sparse} introduced for the image domain tasks, researchers have begun to focus on how to use these Transformer-based models for video object detection. Among them, TransVOD series \cite{zhou2022transvod, he2021end} introduces two key modules to boost the performance: Temporal Query Encoder to fuse object queries and Temporal Deformable Transformer Decoder (TDTD) to obtain current frame detection results. Similarly, PTSEFormer \cite{wang2022ptseformer} proposes the Spatial Transition Awareness Model to fuse the temporal and spatial information for a better prediction. These models still follow the feature-aggregation based prototype, where those of the neighboring frames enhance features of the current frame before being fed into the task heads for the final prediction. Unlike these methods mentioned above, we aggregate the queries of those Transformer-based object detectors for the video tasks, which the current researchers have never investigated. 
\begin{figure*}[!bt]
    \centering
\sbox{\measurebox}{%
  \begin{minipage}[b]{.4\textwidth}
  \subfloat
    []
    {\label{fig:figA}\rotatebox{270}{\includegraphics[width=\textwidth]{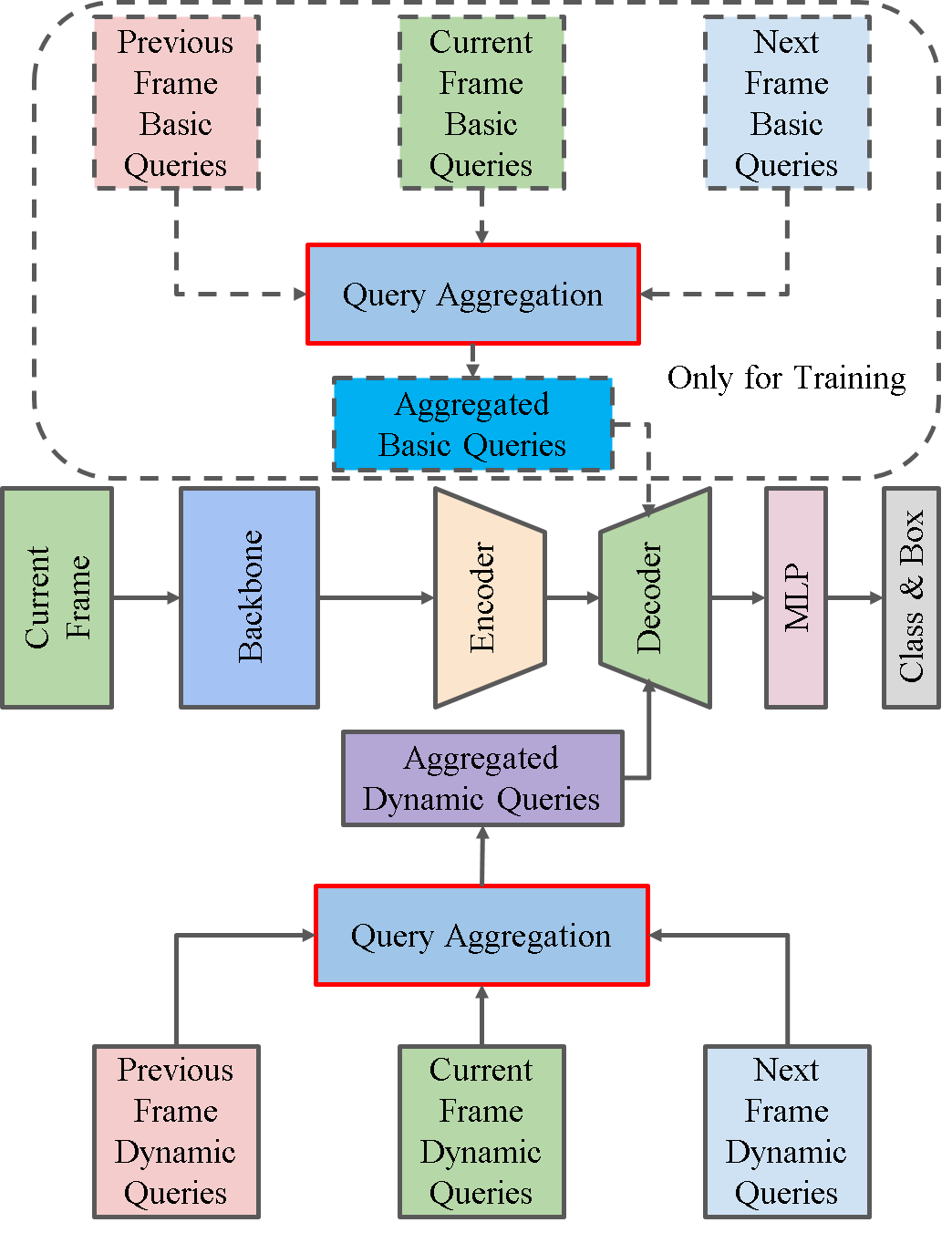}}}
  \end{minipage}}
\usebox{\measurebox}\qquad
\qquad\qquad\qquad
\begin{minipage}[b][\ht\measurebox][s]{.3\textwidth}
\centering
\subfloat
  []
  {\label{fig:figB}\includegraphics[width=0.9\textwidth]{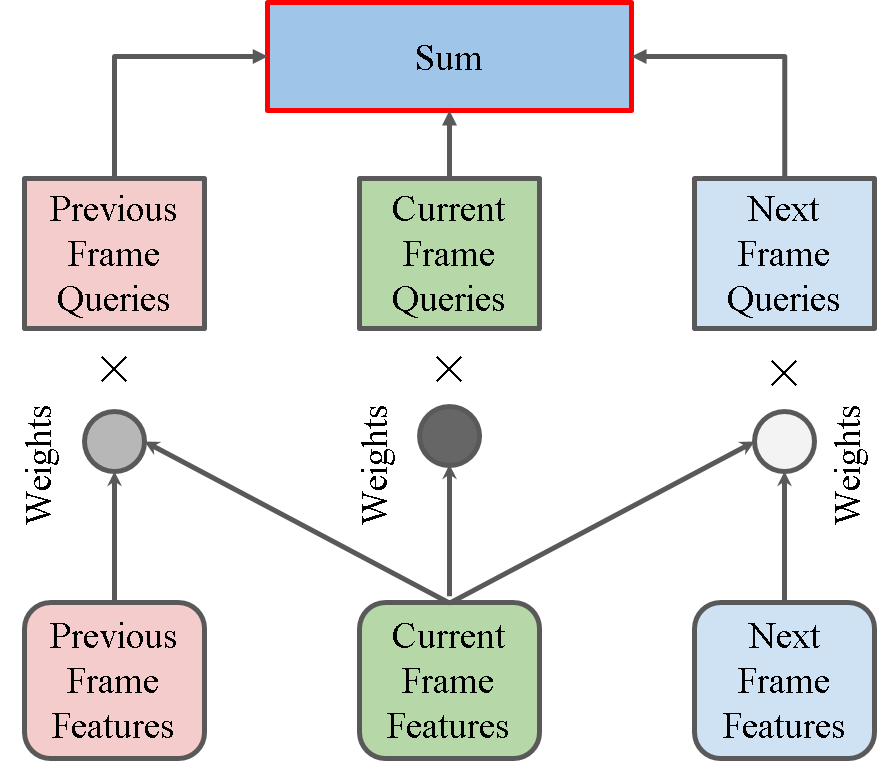}}

\subfloat
  []
  {\label{fig:figC}\includegraphics[width=\textwidth]{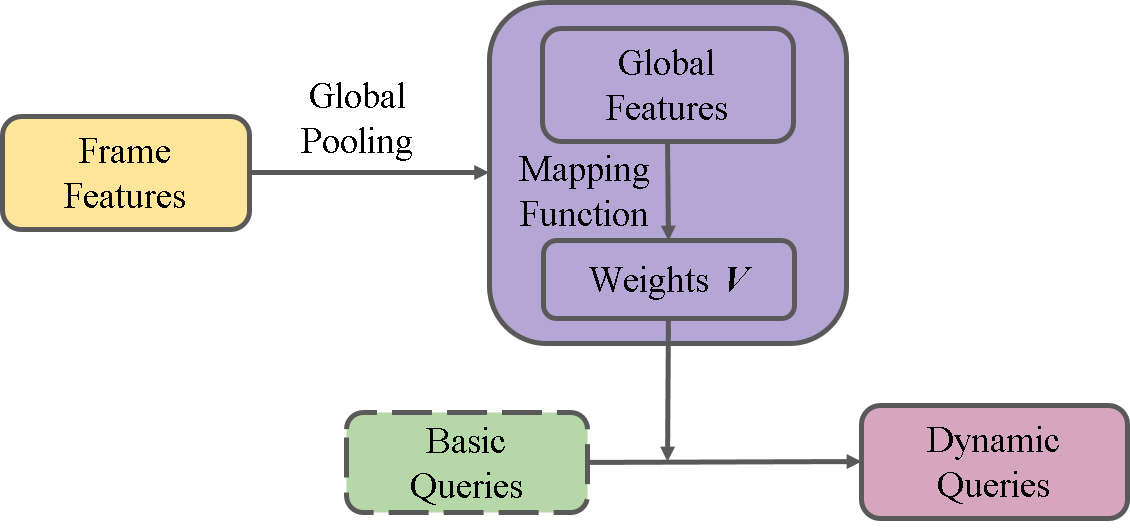}}
\end{minipage}
\caption{Framework of the proposed method. (a) Dynamic query aggregation with both basic queries and dynamic queries. (b) Details of query aggregation. (c) The process to generate the dynamic queries based on the basic queries and input frame features. }
\vspace{-0.2cm}
\label{fig: framework}
\end{figure*}
\noindent\textbf{Dynamic models.} Dynamic networks aim to adjust the inference paths according to the inputs selectively. Slimmable networks \cite{yu2019universally, yu2019autoslim, li2021dynamic} are models which can adaptively select the computational complexity based on the inputs without the need for retraining. In terms of applications for object detection, neural architecture search (NAS) is one of the widely used approaches. NAS-FPN \cite{ghiasi2019fpn} introduces NAS to optimize designing FPNs for object detection according to the input image. In other work like \cite{wang2020fcos, yao2020sm, wang2021fcos}, NAS is applied to the existing image object detectors to improve the performance. Besides NAS, recent works have begun using dynamic models \cite{Gao_2018_CVPR, zhang2020dynamic, zhu2020dynamic, ming2021dynamic, song2020fine,zhang2021dynamic, dai2021dynamic} to improve efficiency dynamic convolutions \cite{chen2020dynamic, dong2021semi, dong2022graph, wang2019dynamic}, dynamic heads \cite{dai2021dynamic2}, and dynamic proposals \cite{cui2022dynamic}, which are introduced to the existing object detectors to balance the performance and inference speed. For video object detection, DFA \cite{cui2022dfa} proposes a model which can dynamically select the frames used to aggregate the features according to the input frames. This work mainly focuses on dynamically aggregating the queries for the Transformer-based object detectors according to the input frames.
% \section{Methodology}
% In this section, we first review the pipeline of the current Transformer-based object detection models. Then, we discuss and compare the existing methods and the models that we are going to propose. Next, we introduce a vanilla version to aggregate queries for video object detection. Finally, we update the vanilla one to a dynamic version, where the queries are aggregated according to the input frames.
\section{Preliminary}
% In this section, we first review the pipeline of the current Transformer-based object detection models. Then, we discuss and compare the existing methods and the models that we are going to propose.

We first review the pipeline of the existing Transformer-based object detectors for videos: Given an input frame $\bm{I}$, the multi-scale features extracted by the backbone like ResNet \cite{he2016deep} are denoted as $\bm{F}$, which are then fed into a Transformer encoder $\mathcal{N}_{enc}$. Next, the outputs of $\mathcal{N}_{enc}$ are fed into a Transformer decoder $\mathcal{N}_{dec}$ together with $n$ randomly initialized queries $\bm{Q}\in\mathbb{R}^{n\times f}$, where $n$ and $f$ denote the number of queries and length of every query respectively. The outputs of the Transformer decoder are then fed into a task head $\mathcal{N}_t$ for the final prediction $\bm{P} =\left\{\left(\bm{b}_{i},\bm{c}_{i}\right), =1,2,\dots,n\right\}$, where $\bm{b}_{i}, \bm{c}_{i}$ represent the corresponding locations and categories of the predicted bounding boxes. The process above can be summarized as follows:
\begin{equation}
    \bm{P} = \mathcal{N}_t\left(\mathcal{N}_{dec}\left(\mathcal{N}_{enc}\left(\bm{F}\right), \bm{Q}\right)\right)
    \label{eq: detr}
\end{equation}

The predictions $\bm{P}$ are then matched with the ground truths $\bm{Y}$ using the Hungarian Algorithm \cite{detr} for bipartite matching, and the final loss is the summation of all the frames, as Equation \ref{eq: bpmatching}.
\begin{equation}
    \mathcal{L} = \sum\mathcal{L}_\text{Hungarian}\left(\bm{P},\bm{Y}\right)
    \label{eq: bpmatching}
\end{equation}

To handle the issue of feature degradation in the video frames,  existing models enhance the representations of different parts in Equation \ref{eq: detr}: Post-processing based models \cite{Kang_2016, han2016seq, kang2017t, lee2016multi} associate the predictions $\bm{P}$ from different frames using temporal hints like optical flows or object tracking to improve the performance of detection at each frame, as Figure \ref{fig: methodCompare}(c). Feature-aggregation based approaches \cite{wu2019sequence, zhu2017flow, wang2018fully, zhou2022transvod, he2021end} enhance the feature representations for object detection by weighted averaging the features $\bm{F}$ neighboring frames, as Figure \ref{fig: methodCompare}(b). Different from the methods mentioned above, we focus on improving the quality of queries $\bm{Q}$ by aggregation, as Figure \ref{fig: methodCompare}(d), which are the unique properties of Transformer-based models compared with the classic approaches.
% \begin{table}[!bt]
%     \centering
%     \begin{tabular}{c|c}
%     \toprule
%      Methods & Target \\
%      \midrule
%     Post-processing\cite{han2016seq, Kang_2016, kang2017t} & $\bm{P}_i$\\
%     Feature-aggregation\cite{zhu2017flow, wu2019sequence, wang2018fully, chen2020memory, gong2021temporal} & $\bm{F}_i$\\
%     Ours & $\bm{Q}_i$\\
%     \bottomrule
%     \end{tabular}
%     \caption{Comparison between the existing models and our proposed module. Target denotes which part in Equation \ref{eq: detr} is improved for video object detection.}
%     \label{tab: comparisonAgg}
% \end{table}
% When it comes to the video object detection, Equation \ref{eq: detr} is updated to be:
% \begin{equation}
%     \bm{P}_i = \mathcal{N}_t\left(\mathcal{N}_{dec}\left(\mathcal{N}_{enc}\left(\bm{F}_i\right), \bm{Q}_i\right)\right)
%     \label{eq: detrVideo},
% \end{equation}
% where $\bm{F}_i, \bm{Q}_i, \bm{P}_i$ represent the corresponding feature map, randomly initialized queries and prediction outputs of the $i^{th}$ frame $\bm{I}_i$.
% 

\section{Query Aggregation}
\label{sec: method}
\subsection{Vanilla Query Aggregation}
In terms of how to aggregate the query $\bm{Q} \in\mathbb{R}^{n\times f}$ of frame $\bm{I}$, a naive vanilla idea is to weighted average the queries $\bm{Q}_{i}$ from the neighboring frame $\bm{I}$, where $\bm{I}_{i}\in\mathcal{N}\left(\bm{I}\right)$\footnote{For simplicity, $\bm{I}$ is also considered to be within the neighboring frames $\mathcal{N}\left(\bm{I}\right)$.} and the size of $\mathcal{N}\left(\bm{I}_i\right)$ is $l$, as Figure \ref{fig: framework} (b). Within one batch, $l\times n$ queries are randomly initialized and shared across the training data during the training process. Within the neighborhood $\mathcal{N}\left(\bm{I}_i\right)$, the queries are different. Therefore, the aggregated query $\Delta\bm{Q}^v \in\mathbb{R}^{n\times f}$ for the current frame $\bm{I}$ is represented as: 
\begin{equation}
    \Delta\bm{Q}^v = \sum_{\forall \bm{I}_{i} \in \mathcal{N}\left(\bm{I}\right)} w_{i}^v\bm{Q}_{i},
    \label{vanillaAgg}
\end{equation}
where $w_{i}^v \in \mathbb{R}$ is the learnable weights for aggregation. A simple idea to generate the learnable weights is based on the cosine similarity of the input frame features, shown as the dots and arrows in Figure \ref{fig: framework}(b). Following the existing feature-aggregation based video object detectors \cite{wu2019sequence, zhu2017flow, wang2018fully}, we generate $w_{i}^v$ according to Equation \ref{eq: similarity}. We will discuss more ways to aggregate the queries in the experiment section.

\begin{equation}
    w_{i}^v = \frac{\alpha(\bm{F})\beta(\bm{F}_{i})}{\lvert\alpha(\bm{F})\rvert\lvert\beta(\bm{F}_{i})\rvert},
    \label{eq: similarity}
\end{equation}
where $\alpha, \beta$ are mapping functions and $\lvert\cdot\rvert$ denotes the normal. The corresponding features of the current frame $\bm{I}$ and its neighbors $\bm{I}_{i}$ are denoted as $\bm{F}$ and $\bm{F}_{i}$. Therefore, the process in Equation \ref{eq: detr} and \ref{eq: bpmatching} are updated as:
\begin{equation}
\begin{aligned}
    \bm{P}^v &= \mathcal{N}_t\left(\mathcal{N}_{dec}\left(\mathcal{N}_{enc}\left(\bm{F}\right), \Delta\bm{Q}^v\right)\right) \\
    \mathcal{L}^v &= \sum\mathcal{L}_\text{Hungarian}\left(\bm{P}^v,\bm{Y}\right),
\end{aligned}
    \label{eq: detrVanilla}
\end{equation}
where $\bm{P}^v$ denotes the prediction results with the aggregated queries $\Delta\bm{Q}^v$. 

% During the training process, we also feed the queries $\bm{Q}_{i}$ without aggregation into the task networks to generate their corresponding prediction $\bm{P}_{i}^v$ and calculate its loss as Equation \ref{eq: detrVanillaNoAgg}:
% \begin{equation}
% \begin{aligned}
%     \bm{P}_{i} &= \mathcal{N}_t\left(\mathcal{N}_{dec}\left(\mathcal{N}_{enc}\left(\bm{F}_i\right), \bm{Q}_{i}\right)\right) \\
%     \mathcal{L} &= \sum_i\mathcal{L}_\text{Hungarian}\left(\Delta\bm{P}_i^v,\bm{Y}_i\right) \\
%     &+ \sum_i\mathcal{L}_\text{Hungarian}\left(\bm{P}_{i}^v, \bm{Y}_i\right)
% \end{aligned}
%     \label{eq: detrVanillaNoAgg}
% \end{equation}
% During the training process, queries $\bm{Q}_{ij}$ from the neighboring frame $\bm{I}_j\in\mathcal{N}\left(\bm{I}_i\right)$ are randomly initialized and interact with each other.

\begin{table*}[!bt]
    \centering
    \begin{tabular}{l|l|l|l|l|l|l}
    \toprule
    Method & AP & AP$_\text{50}$ & AP$_\text{75}$ & AP$_\text{S}$ & AP$_\text{M}$ & AP$_\text{L}$ \\
    \midrule
    \midrule
    \multicolumn{7}{l}{Two-stage Object Detectors}\\
    \midrule
    % Faster R-CNN\cite{ren2015faster} \\
    Faster R-CNN\cite{ren2015faster} + DFF \cite{zhu2017deep}  & 42.7 & 70.3 & 45.7 & 5.0 & 17.6 & 48.7 \\
    Faster R-CNN\cite{ren2015faster} + FGFA \cite{zhu2017flow} & 47.1 & 74.7 & 52.0 & 5.9 & 22.2 & 53.1\\
    Faster R-CNN\cite{ren2015faster} + SELSA \cite{wu2019sequence} & 48.7 & 78.4 & 53.1 & 8.5 & 26.3 & 54.5\\
    Faster R-CNN\cite{ren2015faster} + Temporal RoI Align \cite{gong2021temporal} & 48.5 & 79.8 & 52.3 & 7.2 & 26.5 & 54.4\\
    % Faster RCNN + MEGA \cite{chen2020memory} & \\
    \midrule
    \midrule
    \multicolumn{7}{l}{DETR-based Object Detectors}\\
    \midrule
    SMCA-DETR\cite{gao2021fast} & 53.5 & 74.2 & 59.6 & 7.6 & 25.7 & 60.5\\
    SMCA-DETR\cite{gao2021fast} + TransVOD\cite{zhou2022transvod} & 54.6$_{\uparrow1.1}$ & 78.5$_{\uparrow4.3}$ & 61.0$_{\uparrow1.4}$ & 9.1$_{\uparrow1.5}$ & 27.0$_{\uparrow1.3}$ & 61.4$_{\uparrow0.9}$ \\
    SMCA-DETR\cite{gao2021fast} + Ours & 55.8$_{\uparrow2.3}$ & 79.1$_{\uparrow4.9}$ & 62.7$_{\uparrow3.3}$ & 9.3$_{\uparrow1.7}$ & 28.7$_{\uparrow3.0}$ & 62.6$_{\uparrow2.1}$\\
    \midrule
    Conditional-DETR\cite{Meng_2021_ICCV} & 53.7 & 74.7 & 60.1 & 7.7 & 25.9 & 60.6\\
    Conditional-DETR\cite{Meng_2021_ICCV} + TransVOD\cite{zhou2022transvod} & 54.8$_{\uparrow1.1}$ & 78.6$_{\uparrow3.9}$ & 61.3$_{\uparrow1.2}$ & 9.6$_{\uparrow1.9}$ & 27.5$_{\uparrow1.6}$ & 61.9$_{\uparrow1.3}$ \\
    Conditional-DETR\cite{Meng_2021_ICCV} + Ours & 56.1$_{\uparrow2.4}$ & 79.2$_{\uparrow4.5}$ & 63.0$_{\uparrow2.9}$ & 8.8$_{\uparrow1.1}$ & 29.0$_{\uparrow3.1}$ & 63.0$_{\uparrow2.4}$\\
    \midrule
    DAB-DETR\cite{liu2022dab} & 54.2 & 75.3 & 61.3 & 8.9 & 26.8 & 61.2\\
    DAB-DETR\cite{liu2022dab} + TransVOD\cite{zhou2022transvod} & 56.4$_{\uparrow2.2}$ & 77.2$_{\uparrow1.9}$ & 63.7$_{\uparrow2.4}$ & 10.1$_{\uparrow1.2}$ & 28.9$_{\uparrow2.1}$ & 63.5$_{\uparrow2.3}$\\
    DAB-DETR\cite{liu2022dab} + Ours & 58.0$_{\uparrow3.8}$ & 79.0$_{\uparrow3.7}$ & 65.5$_{\uparrow4.2}$ & 12.0$_{\uparrow3.1}$ & 30.1$_{\uparrow3.1}$ & 65.1$_{\uparrow3.9}$ \\
    \midrule
    % Deformable-DETR\cite{zhu2021deformable} & 57.1 & 77.5 & 63.5 & 11.2 & 30.0 & 64.0\\
    Deformable-DETR\cite{zhu2021deformable} & 55.4 & 76.2 & 62.2 & 10.5 & 27.5 & 62.3\\
    Deformable-DETR\cite{zhu2021deformable} + TransVOD \cite{zhou2022transvod} & 58.1$_{\uparrow2.7}$ & 79.1$_{\uparrow2.9}$ & 64.7$_{\uparrow2.5}$ & 11.0$_{\uparrow0.5}$ & 29.9$_{\uparrow2.4}$ & 65.2$_{\uparrow2.9}$\\
    Deformable-DETR\cite{zhu2021deformable} + Ours &  60.1$_{\uparrow4.7}$ & 81.7$_{\uparrow5.5}$ & 66.9$_{\uparrow4.7}$ & 13.2$_{\uparrow2.7}$ & 33.1$_{\uparrow5.6}$ & 66.9$_{\uparrow4.6}$\\
    \midrule
    % DAB-Deformable-DETR\cite{liu2022dab}\\
    % DAB-Deformable-DETR\cite{liu2022dab} + TransVOD\cite{zhou2022transvod} \\
    % DAB-Deformable-DETR\cite{liu2022dab} + Ours\\
    % \midrule
    \bottomrule
    \end{tabular}
    \caption{Performance comparison with the recent state-of-the-art video object detection approaches on ImageNet VID validation set \cite{russakovsky2015imagenet}. The AP$_\text{50}$ here is the mAP evaluation metric in most of the existing works like TransVOD \cite{zhou2022transvod}  and PTSEFormer \cite{wang2022ptseformer}.}
        \vspace{-0.3cm}
    \label{tab: mainResult}
\end{table*}
\subsection{Dynamic Query Aggregation}
The vanilla query aggregation module has an issue that these neighboring queries $\bm{Q}_{i}$ are randomly initialized, which are not related to their corresponding frames $\bm{I}_{i}$. Therefore, the neighboring queries $\bm{Q}_{i}$ do not provide enough temporal or semantic information to eliminate the feature degradation issues caused by fast motion. Though the weights $w_{i}^v$ used for aggregation are related to the features $\bm{F}_{i}$ and $\bm{F}$, there are not enough constraints on the quantities of those randomly initialized queries $\bm{Q}_{i}$. 

Therefore, we propose to update the vanilla query aggregation module to a dynamic version, which adds constraints to the queries and can adjust the weights according to the neighboring frames. For implementation, the simple idea is to generate the queries $\bm{Q}_{i}$ directly from the features $\bm{F}_{i}$ of the input frame. However, experiments show us that this way is challenging to train and always gets a worse performance. Unlike the naive idea mentioned above, we generate the new queries adaptive to the input frame from the randomly initialized queries. We first define two kinds of query vectors, shown with dashed and solid lines in Figure \ref{fig: framework} (a) and (c): basic queries $\bm{Q}_{i}^b \in\mathbb{R}^{n\times f}$ and dynamic queries $\bm{Q}_{i}^d \in \mathbb{R}^{m\times f}$, where $n = rm$ and $r$ is set to be $4$ by default. During the training and inference processes, we generate the dynamic queries from the basic queries according to the features $\bm{F}_i, \bm{F}$ of the input frames as:
\begin{equation}
    \bm{Q}_i^d = \mathcal{M}\left(\bm{Q}_{i}^b, \bm{F}_i, \bm{F}_{i}\right), 
    \label{eq: modulation}
\end{equation}
where $\mathcal{M}\left(\cdot\right)$ is the mapping function to build the relationship of the basic query $\bm{Q}_{i}^b$ and the dynamic one $\bm{Q}_{i}^d$ according to $\bm{F}$ and $\bm{F}_{i}$. Here we give a default example of the implementation of $\mathcal{M}\left(\cdot\right)$ and will analyze its design in the experiment section in detail. We first divide the basic queries $\bm{Q}_i^b$ into $m$ groups, where each group has $r$ queries. Then, for each group, we use the same weights $\bm{V} = \left\{v_j, j = 1, 2, \dots, r\right\}$, $\bm{V}\in\mathbb{R}^{r\times m}$ to weighted average the queries in the current group, as Equation \ref{eq: weighted}.

\begin{equation}
    \bm{Q}_{i}^d = \sum_{j=1}^r v_{j}\bm{Q}_{ij}^b,
    \label{eq: weighted}
\end{equation}
where $\bm{Q}_{ij}^b$ denotes the $j$-th basic query in the current group of $\bm{Q}_i^b$. To build the relationship between the dynamic queries $\bm{Q}_i^d$ and their corresponding frame $\bm{I}_i$, we generate the weights $\bm{V}$ using the global features of $\bm{i}_i$, as Figure \ref{fig: framework}(c), denoted as:
\begin{equation}
    \bm{V} = \mathcal{G}\left(\mathcal{A}\left(\bm{F}_i\right)\right),
    \label{eq: weightGeneration}
\end{equation}
where $\mathcal{A}$ is a global pooling operation to reduce the feature resolution and generate the global-level features, and $\mathcal{G}$ is a mapping function to project the pooled features to the dimension of $r\times m$. Therefore, the process of aggregating the queries dynamically based on the features of the input frames can be updated as follows: 
\begin{equation}
\begin{aligned}
    \Delta\bm{Q}_i^d &= \sum_{\forall \bm{I}_i \in \mathcal{N}\left(\bm{I}\right)} w_{i}^d\bm{Q}_{i}^d \\ 
    % &= \sum_{\forall \bm{I}_j \in \mathcal{N}\left(\bm{I}_i\right)} w_{ij}^d\mathcal{M}\left(\bm{Q}_{ij}, \bm{F}_i, \bm{F}_j\right),
\end{aligned}
    \label{dynamicAgg}
\end{equation}

% Then, the predictions based on the dynamic query aggregation is updated to be:
% \begin{equation}

%     \label{eq: detrDynamic}
% \end{equation}

During the training, as shown in Figure \ref{fig: framework}(a), we aggregate both the dynamic queries $\bm{Q}_i^d$ and basic queries $\bm{Q}_i^b$ with the same weights $w_i^d$ and generate the corresponding prediction $\bm{P}^d$ and $\bm{P}^b$, as Equation \ref{eq: detrDynamic}.  
\begin{equation}
\begin{aligned}
    \Delta\bm{Q}^b &= \sum_{\forall \bm{I}_i \in \mathcal{N}\left(\bm{I}\right)} w_{i}^d\bm{Q}_{i}^b \\ 
    \bm{P}^b &= \mathcal{N}_t\left(\mathcal{N}_{dec}\left(\mathcal{N}_{enc}\left(\bm{F}\right), \Delta\bm{Q}^b\right)\right) \\
    \bm{P}^d &= \mathcal{N}_t\left(\mathcal{N}_{dec}\left(\mathcal{N}_{enc}\left(\bm{F}\right), \Delta\bm{Q}^d\right)\right)
\end{aligned}
    \label{eq: detrDynamic}
\end{equation}

We calculate the bipartite matching losses for both $\bm{P}^b$ and $\bm{P}^d$ and use a hyperparameter $\gamma$ to balance their influence as Equation \ref{eq: bpmatchingDynamic}.
\begin{equation}
    \mathcal{L} = \sum\mathcal{L}_\text{Hungarian}\left(\bm{P}^d,\bm{Y}\right) + \gamma \sum\mathcal{L}_\text{Hungarian}\left(\bm{P}^b,\bm{Y}\right)
    \label{eq: bpmatchingDynamic}
\end{equation}
\begin{table}[!bt]
    \centering
    \begin{tabular}{c|c|c|c|c}
    \toprule
    Model & VQA &  DQA & Loss & mAP \\
    \midrule
    A & & & & 55.4\\
    B & $\checkmark$ & & & 56.7\\
    C & & $\checkmark$ & & 58.5\\
    D & & & $\checkmark$ & 58.9\\
    E & & $\checkmark$ & $\checkmark$ & 60.1\\
    \bottomrule
    \end{tabular}
    \caption{Analysis of the proposed modules.}
        \vspace{-0.4cm}
    \label{tab: component}
\end{table}

During the inference time, we only use the dynamic queries $\bm{Q}_i^d$ and their corresponding predictions $\bm{P}^d$ as the final outputs, which introduce only a little extra computational complexity to the original models. We will discuss and analyze the model complexity in the experiment part.
% \begin{equation}
%     \bm{Q}_{ij}^d = \mathcal{A}\left(\bm{Q}_{ij}\right)
% \end{equation}
% \clearpage
\section{Experiments}
\subsection{Experimental Setup}
% \noindent\textbf{Dataset.} 
We evaluate our proposed methods on the ImageNet VID benchmark \cite{russakovsky2015imagenet} with the recent state-of-the-art Transformer-based object detection models \cite{gao2021fast, zhu2021deformable, liu2022dab, Meng_2021_ICCV}. Following the pipeline of TransVOD \cite{zhou2022transvod, he2021end}, we first pretrain our models on the MS COCO \cite{lin2014microsoft} and then fine-tune on the combination of ImageNet VID and DET datasets. 
% During the fine tune process, we first remove the query aggregation module to refine the classification networks from the MS COCO to the ImageNet VID benchmark. Then, we set the number of neighboring frames for query aggregation to be $14$ by default to train the query aggregation module. During this process, the modules besides the query aggregation parts are frozen. For the basic queries $\bm{Q}_{ij}^b$, they are randomly initialized while the dynamic queries $\bm{Q}_{ij}^d$ are generated based on $\bm{Q}_{ij}^b$. Therefore, the basic queries $\bm{Q}_{ij}^b$ are shared regardless of the inputs while $\bm{Q}_{ij}^d$ are not. In other words, the $k$-th basic query for the $k$-th frame used for aggregation is always the same. 
All the models are trained on $8$ Tesla A100 GPUs, and during the training and inference processes, $14$ neighboring frames are used for aggregation. 

% \noindent\textbf{Implementation details.} During the training process, following TransVOD \cite{zhou2022transvod, he2021end}, we first fine tune the models pretrained on the MS COCO \cite{lin2015microsoft} benchmark for the image object detection. 

% For the vanilla query aggregation module, $\alpha, \beta$ are implemented as two two-layer MLPs. 

\subsection{Main Results}
In this section, we conduct experiments on the dynamic query aggregation modules with the current Transformer-based object detectors on the ImageNet VID benchmark \cite{russakovsky2015imagenet}. The experiments of vanilla query aggregation will be provided in the following section. For a fair comparison, we use the same experimental setups and compare these models with and without our proposed modules integrated. The default backbone is ResNet-50 \cite{he2016deep}. We summarize the results in Table \ref{tab: mainResult}. 
% Visualization examples are provided in the supplementary materials.

For most of the DETR-based object detectors, the performance can be improved by at least $2.4\%$ on the metric of mAP compared with those not integrated with our proposed modules. The performance is even better than those integrated with TransVOD \cite{zhou2022transvod} by a large margin, which validates the effectiveness of our dynamic query aggregation modules. When considering the objects' sizes, we notice that the performance of large or medium objects is much better than that of small objects. We argue that this is because the process to generate the dynamic queries $\bm{Q}_{i}^d$ is only based on the global features, which lack enough information for the small objects. 
% \begin{figure*}[!bt]
%     \centering
%   \includegraphics[width=16cm]{img/obj.png}
%     \caption{Caption}
%     \label{fig: visualization}
% \end{figure*}
\subsection{Model Analysis}
We conduct experiments with Deformable-DETR \cite{zhu2021deformable} on the ImageNet VID benchmark \cite{russakovsky2015imagenet} in this section to study the design of our proposed modules, mainly the dynamic query aggregation module. 
% More experiment results are provided in the supplementary materials.

\noindent\textbf{Analysis of $\mathcal{M}$.} We conduct experiments to analyze the design of $\mathcal{M}$, as Table \ref{tab: analysisM}. Model A is our default setting where only one $\bm{V}$ is generated, and $r$ is set to be $4$. Model B and C increase the number of weight matrix $\bm{V}$ to be $2$ and $4$. In detail, multiple outputs are provided in each group of $\bm{Q}_i^b$ instead of generating only one $\bm{Q}_i^d$. Model D replaces the $\mathcal{G}$ and $\mathcal{A}$ operations in Equation \ref{eq: weightGeneration} to MLPs, and Model E generates $\bm{Q}_i^d$ from $\bm{Q}_i^b$ and $\bm{F}_i$ using multi-head cross attention. Models F and G follow model A but have different ways of grouping the basic queries. Model F shuffles the queries before grouping, and model G randomly clusters non-consecutive queries into a group. From the table, we notice that by increasing the numbers of $\bm{V}$, there is a slight improvement in the performance at the sacrifice of more extra parameters. For models D and E, the performance drops a little bit. We argue that this is because local features from $\bm{F}_i$ will bring misleading information and make the model difficult to optimize. For models F and G, there is not much difference with model A because the basic queries $\bm{Q}_i^b$ are randomly initialized, bringing robustness to our models.
\begin{table}[! bt]
    \centering
    \begin{tabular}{c|c|c|c|c}
    \toprule
    Model & mAP & AP$_\text{50}$ & AP$_\text{75}$ & Extra Param\\
    \midrule
    A & 60.1 & 81.7 & 66.9 & 37K\\
    B & 60.3 & 82.0 & 67.0 & 74K\\
    C & 60.5 & 82.3 & 67.2 & 147K\\
    D & 59.3 & 80.7 & 65.8 & 49K\\
    E & 58.2 & 79.3 & 64.9 & 26K\\
    F & 60.0 & 81.7 & 66.8 & 37K\\
    G & 60.1 & 81.6 & 67.0 & 37K\\
    \bottomrule
    \end{tabular}
        \vspace{-0.2cm}
    \caption{Analysis of different designs on $\mathcal{M}.$}
    \label{tab: analysisM}
\end{table}
\noindent\textbf{Analysis of aggregation process.} In section \ref{sec: method}, we provide a way to implement the aggregation process based on cosine similarity as the existing works \cite{zhu2017flow, chen2020memory, wang2018fully, wu2019sequence}. Here, we analyze different ways to aggregate the queries as Table \ref{tab: ablationAgg}. Besides cosine similarity, we use simple learnable networks as TF-Blender \cite{cui2021tf} and Transformer as TransVOD \cite{zhou2022transvod, he2021end}. For a fair comparison, all the models are trained with $14$ neighboring frames and tested on the ImageNet VID benchmark \texttt{val} split. For the implementation details, we use the same structures as TF-Blender \cite{cui2021tf} and TransVOD \cite{zhou2022transvod, he2021end}. The table shows that cosine similarity based aggregation has the worst performance compared to learnable simple networks and Transformers. By default, we use the Transformer to aggregate the queries in our work. 

\begin{table}[!bt]
    \centering
    \begin{tabular}{c|c|c|c}
    \toprule
    Aggregation Methods & mAP & AP$_\text{50}$ & AP$_\text{75}$  \\
    \midrule
    Cosine Similarity & 58.7 & 79.3 & 64.9\\
    Simple Networks & 59.5 & 80.2 & 65.3\\
    Transformer & 60.1 & 81.7 & 66.9\\
    \bottomrule
    \end{tabular}
    \caption{Analysis of different ways to aggregate queries.}
    \vspace{-0.4cm}
    \label{tab: ablationAgg}
\end{table}

\noindent\textbf{Analysis of each component.} We conduct experiments to study the effects of each proposed component. We denote the original Deformable-DETR \cite{zhu2021deformable} with ResNet-50 as the backbone of model A. Then, we introduce the vanilla query aggregation modules with $14$ neighboring frames to the original Deformable-DETR to get model B. For model C, we change the vanilla query aggregation module to the dynamic version without the extra loss. To validate the effectiveness of our proposed modules, we conduct experiments on mode D, where we use two groups of randomly initialized $\bm{Q}_{i}^b$ and $\bm{Q}_{i}^d$. We do not build a relationship between them. Therefore, there is only a loss for these two groups of queries but no relation to the input frames. Finally, we integrate the dynamic query aggregation module and the extra loss to get model E. The results are summarized in Table \ref{tab: component}.

From the table, we notice that by only using the vanilla query aggregation module, the performance can be improved by $1.3\%$ compared with the original Deformable-DETR without aggregation. However, this is worse than the original TransVOD \cite{zhou2022transvod} model. Updating the vanilla query aggregation module to the dynamic version increases the performance by $1.8\%$, which is better than the original TransVOD. We argue that by introducing the contents of the inputs into the queries, the performance can be improved by a large margin. Model D removes the relations between the basic and dynamic queries but leaves both losses for them. It also performs well, though different than model E, which contains both the dynamic query aggregation module and the extra losses. From the experiments, we notice that by either introducing the dynamic query aggregation or adding two separate groups of randomly initialized queries, though they are not related to the inputs, the performance can also be improved. By combining these two modules, our proposed methods achieve the best performance. 

\begin{figure}[!bt]
    \centering
    \begin{subfigure}[b]{0.4\textwidth}
         \centering
         \includegraphics[width=6cm]{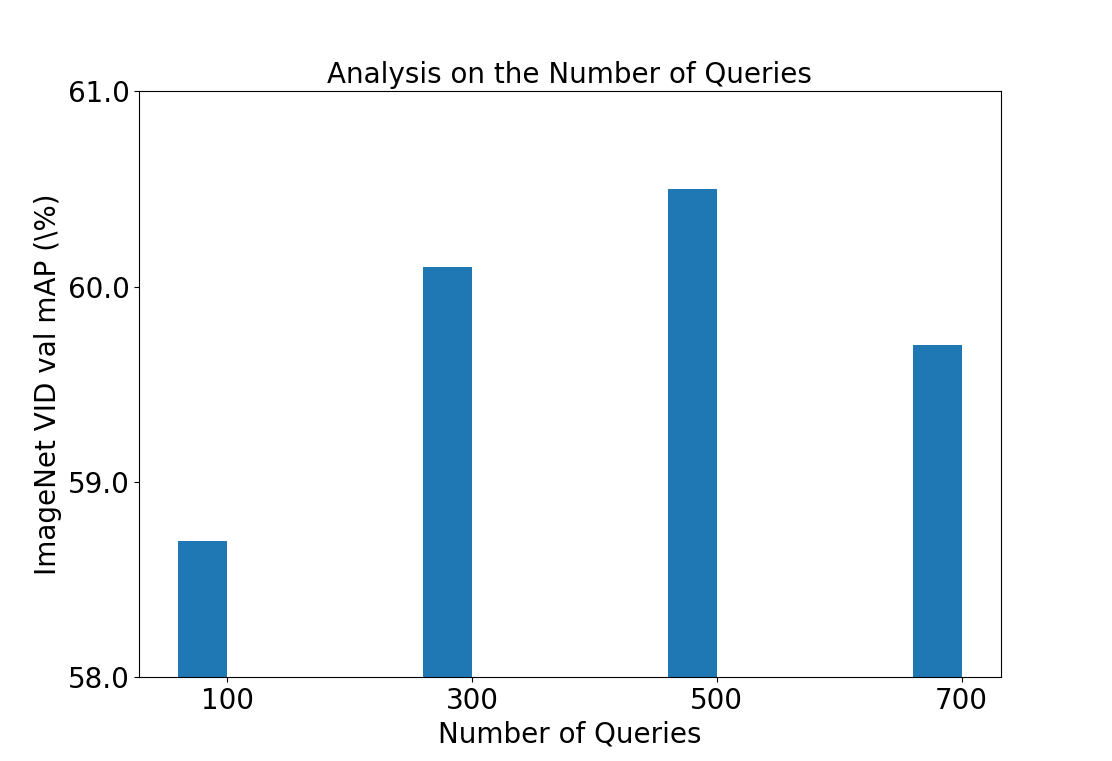}
         \caption{Analysis of the different number of queries}
     \end{subfigure}
     \begin{subfigure}[b]{0.4\textwidth}
         \centering
         \includegraphics[width=6cm]{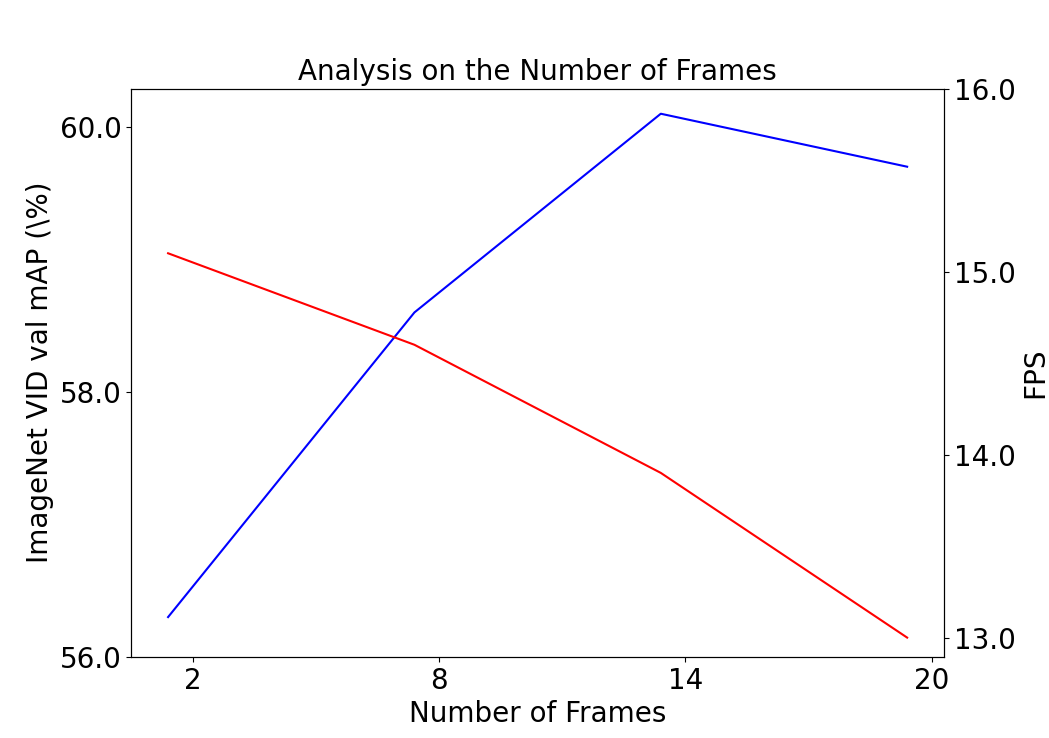}
         \caption{Analysis of the different number of frames}
     \end{subfigure}
    \caption{Model analysis on the number of queries and frames used for aggregation. The default model is Deformable-DETR\cite{zhu2021deformable} with ResNet-50 as the backbone. }
    \label{fig: ablation}
\end{figure}

\noindent\textbf{Analysis of the number of queries.} We conduct experiments with dynamic query aggregation on the number of queries as Figure \ref{fig: ablation} (a). From the figure, by increasing the number of queries, the performance of the video object detection will be improved accordingly. However, when the number of queries is more than $500$, the performance begins to be saturated. We argue that this is because there are enough queries to represent the objects in the input frames, and some are redundant.

\begin{table}[!bt]
    \centering
    \begin{tabular}{c|c|c|c}
    \toprule
    Model & $r$ & mAP & AP$_\text{50}$ \\
    \midrule
    Deformable-DETR & - & 55.4 & 76.2\\
    \midrule
    \multirow{4}{*}{Deformable-DETR + Ours} & 1 & 58.7 & 80.9\\
    & 2 & 59.4 & 81.2\\
    & 4 & 60.1 & 81.7\\
    & 8 & 59.7 & 81.5\\
    \bottomrule
    \end{tabular}
    \caption{Analysis of the effect of $r$. Experiments are conducted with Deformable-DETR \cite{zhu2021deformable} using ResNet-50 as the backbone.}
        \vspace{-0.4cm}
    \label{tab: ablationR}
\end{table}

\noindent\textbf{Analysis of model complexity.} Here, we analyze the model complexity of our proposed modules to the existing object detectors. These methods mainly have four computational loads: 1. Feature extraction network from the backbone $\mathcal{N}_{ex}$; 2. Transformer encoder networks $\mathcal{N}_{enc}$; 3. Transformer decoder networks $\mathcal{N}_{dec}$; 4. task network $\mathcal{N}_{tk}$. Therefore, the total computational complexity is:
\begin{equation}
\mathcal{O}\left(\mathcal{N}_{tk}\right) + \mathcal{O}\left(\mathcal{N}_{dec}\right) + 
\mathcal{O}\left(\mathcal{N}_{enc}\right) + 
\mathcal{O}\left(\mathcal{N}_{ex}\right)
\end{equation}
In our proposed models, we introduce a tiny network $\mathcal{M}$ to generate the dynamic queries $\bm{Q}_{i}^d$. Therefore, during the training process, the computational complexity of our model is defined as:
\begin{equation}
\begin{aligned}
& \mathcal{O}\left(\mathcal{N}_{tk}\right) + 
\mathcal{O}\left(\mathcal{N}_{dec}\right) * (r+1) \\
&+
\mathcal{O}\left(\mathcal{N}_{enc}\right) +
\mathcal{O}\left(\mathcal{M}\right) + \sum\mathcal{O}\left(\mathcal{N}_{ex}\right) ,
\end{aligned}
\end{equation}
where $r+1$ represents the extra loss calculated for aggregation (both $\bm{Q}_{i}^d$ and $\bm{Q}_{i}^b$). Typically, in the Transformer-based object detectors, $\mathcal{O}\left(\mathcal{M}\right) \ll \mathcal{O}\left(\mathcal{N}_{ex}\right) \approx \mathcal{O}\left(\mathcal{N}_{dec}\right) < \mathcal{O}\left(\mathcal{N}_{enc}\right)$. Therefore, our model does not increase too many computational loads to the existing models. When it comes to the inference process, the computational complexity is updated to be:
\begin{equation}\small
\mathcal{O}\left(\mathcal{N}_{tk}\right) + 
\mathcal{O}\left(\mathcal{N}_{dec}\right) +
\mathcal{O}\left(\mathcal{N}_{enc}\right) +
\mathcal{O}\left(\mathcal{M}\right) + \sum\mathcal{O}\left(\mathcal{N}_{ex}\right),
\end{equation}
since only the dynamic queries $\bm{Q}_{i}^d$ are taken into account. The increasing computational load is affordable since the impact of $\mathcal{O}\left(\mathcal{M}\right)+ \sum\mathcal{O}\left(\mathcal{N}_{ex}\right)$ is negligible compared to those of the other networks structures like $\mathcal{O}\left(\mathcal{N}_{dec}\right)$.

\noindent\textbf{Analysis of the number of frames used for aggregation.} Continuing the above part, we conduct experiments to analyze the effect of the number of frames used for aggregation here, as Figure \ref{fig: ablation} (b). The performance will increase by increasing the number of frames used for aggregation. However, when the number of frames is more than $16$, the performance begins to be saturated. Therefore, we use $14$ as the default number of frames for aggregation.

\noindent\textbf{Analysis of $r$.} We conduct experiments to analyze the effect of $r$ on the final performance. By default, $m$ is set to be $300$ for the dynamic queries so that the number of queries will not affect the final performance during the inference time. We change the value of $r$ to use different numbers of basic queries $\bm{Q}_{i}^b$ to generate the same number of dynamic queries $\bm{Q}_{i}^d$, and the results are summarized in Table \ref{tab: ablationR}. The table shows that the performance is the worst when $r$ is set to $1$. We argue that a limited number of basic queries are not enough to generate the dynamic queries adaptive to the input frames. However, the performance will be saturated when $r$ is set to be $8$. We think this is because there are enough and even redundant basic queries to generate the dynamic ones.

% \begin{table}[!bt]
%     \centering
%     \begin{tabular}{c|c|c|c|c|c|c}
%     \toprule
%     $\gamma$ & mAP & AP$_\text{50}$ & AP$_\text{75}$ & AP$_\text{S}$ & AP$_\text{M}$ & AP$_\text{L}$ \\
%     \midrule
%     0.0 &  \\
%     0.5 & \\
%     1.0 & \\
%     \bottomrule
%     \end{tabular}
%     \caption{Caption}
%     \label{tab: extraLoss}
% \end{table}

\begin{figure}[!bt]
    \centering
    \includegraphics[width=7.6cm,trim={4cm 6cm 1cm 4cm},clip]{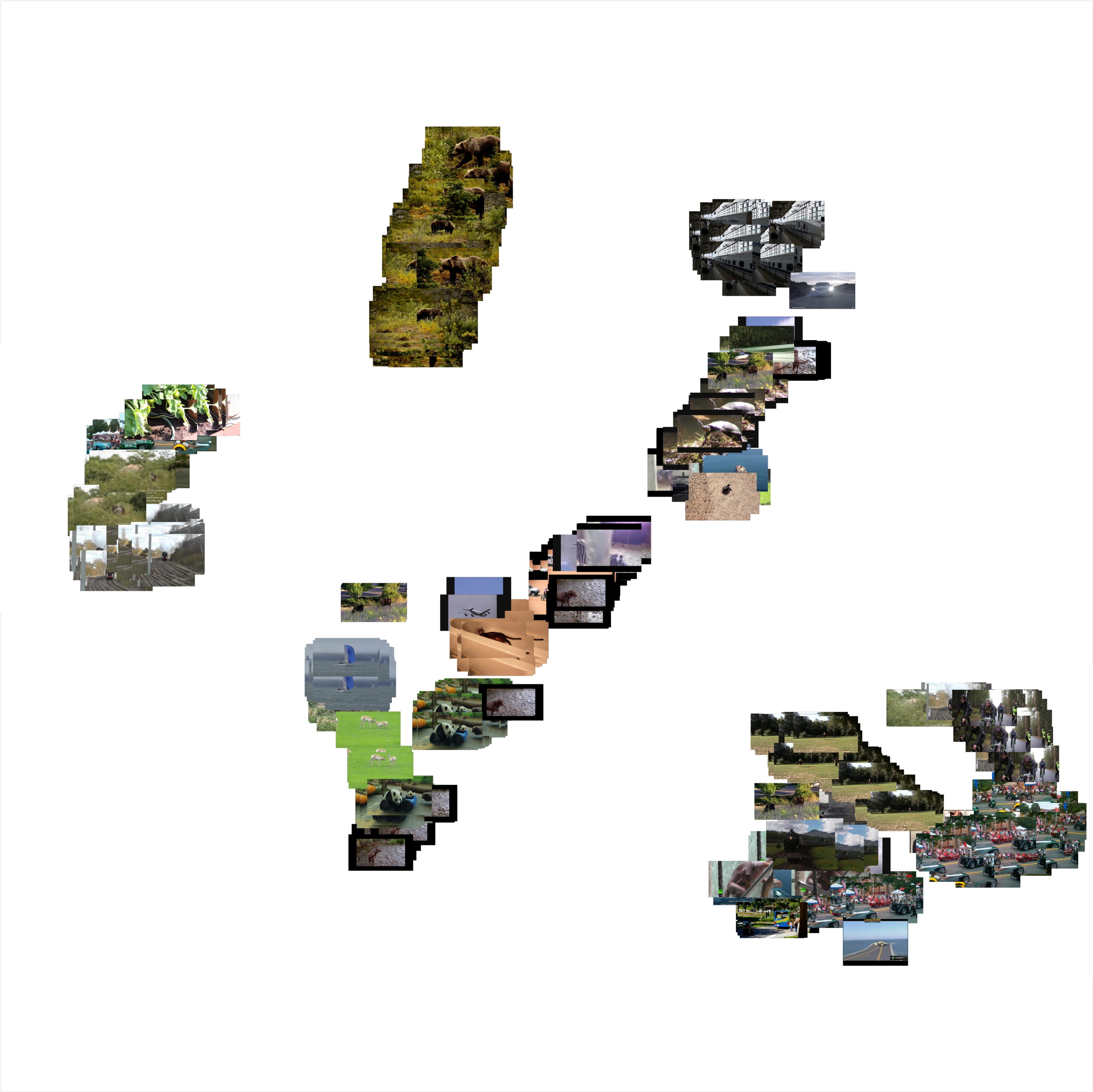}
    \caption{Visualization of dynamic queries with TSNE \cite{van2008visualizing}. Please zoom in for better visualization.}
    \label{fig: tSNE}
    \vspace{-0.4cm}
\end{figure}

\noindent\textbf{Analysis of $\bm{Q}_{i}^b$ and $\bm{Q}_{i}^d$.} To better understand our proposed method, we analyze and visualize the queries from the original models and the dynamic queries from our models. We choose $100$ video clips from the ImageNet VID benchmark \cite{russakovsky2015imagenet} and sample $14$ frames from each video. We generate the corresponding queries based on the input frames and visualize them using the TSNE \cite{van2008visualizing} as Figure \ref{fig: tSNE}. For the original model, the queries are always the same regardless of the input frames\footnote{It is the same with our basic queries. Since all the input frames share the same queries, we do not visualize them.}. Regarding our dynamic queries, as Figure \ref{fig: tSNE}, queries within the same video clips share similar representations, like the clusters on the top and left, which are better and easier to improve the performance of video object detection.
\section{Conclusion}
In this paper, we discuss the unique property of the existing Transformer-based image object detectors and introduce a plug-and-play module designed specifically for these models for the video domain tasks. We first introduce a vanilla version to aggregate the queries for the decoders of the existing Transformer-based models to improve the performance of video object detection. Then, we extend the vanilla query aggregation module to a dynamic version which builds the relationships between the queries and the features of the input frame. Extensive experiments demonstrate that, when integrated with our proposed modules, the current state-of-the-art Transformer-based image object detectors can perform much better on the video object detection task. We believe our proposed modules can bring some light to the Transformer-based models for the video tasks.

%%%%%%%%% REFERENCES
{\small
\bibliographystyle{ieee_fullname}
\bibliography{ref}

\begin{thebibliography}{10}\itemsep=-1pt

\bibitem{belhassen2019improving}
Hatem Belhassen, Heng Zhang, Virginie Fresse, and El-Bay Bourennane.
\newblock Improving video object detection by seq-bbox matching.
\newblock In {\em VISIGRAPP}, 2019.

\bibitem{bodla2017soft}
Navaneeth Bodla, Bharat Singh, Rama Chellappa, and Larry~S Davis.
\newblock Soft-nms--improving object detection with one line of code.
\newblock In {\em ICCV}, 2017.

\bibitem{Cai_2019}
Zhaowei Cai and Nuno Vasconcelos.
\newblock Cascade r-cnn: High quality object detection and instance
  segmentation.
\newblock {\em IEEE Transactions on Pattern Analysis and Machine Intelligence},
  page 1–1, 2019.

\bibitem{detr}
Nicolas Carion, Francisco Massa, Gabriel Synnaeve, Nicolas Usunier, Alexander
  Kirillov, and Sergey Zagoruyko.
\newblock End-to-end object detection with transformers.
\newblock In {\em ECCV}, 2020.

\bibitem{chen2022group}
Qiang Chen, Xiaokang Chen, Gang Zeng, and Jingdong Wang.
\newblock Group detr: Fast training convergence with decoupled one-to-many
  label assignment.
\newblock {\em arXiv preprint arXiv:2207.13085}, 2022.

\bibitem{chen2020memory}
Yihong Chen, Yue Cao, Han Hu, and Liwei Wang.
\newblock Memory enhanced global-local aggregation for video object detection.
\newblock In {\em CVPR}, 2020.

\bibitem{chen2020dynamic}
Yinpeng Chen, Xiyang Dai, Mengchen Liu, Dongdong Chen, Lu Yuan, and Zicheng
  Liu.
\newblock Dynamic convolution: Attention over convolution kernels.
\newblock In {\em CVPR}, 2020.

\bibitem{cui2022dfa}
Yiming Cui.
\newblock Dfa: Dynamic feature aggregation for efficient video object
  detection.
\newblock {\em arXiv preprint arXiv:2210.00588}, 2022.

\bibitem{cui2021tf}
Yiming Cui, Liqi Yan, Zhiwen Cao, and Dongfang Liu.
\newblock Tf-blender: Temporal feature blender for video object detection.
\newblock In {\em ICCV}, 2021.

\bibitem{cui2022dynamic}
Yiming Cui, Linjie Yang, and Ding Liu.
\newblock Dynamic proposals for efficient object detection.
\newblock {\em arXiv preprint arXiv:2207.05252}, 2022.

\bibitem{dai2016r}
Jifeng Dai, Yi Li, Kaiming He, and Jian Sun.
\newblock R-fcn: Object detection via region-based fully convolutional
  networks.
\newblock {\em NeurIPS}, 29, 2016.

\bibitem{dai2021dynamic2}
Xiyang Dai, Yinpeng Chen, Bin Xiao, Dongdong Chen, Mengchen Liu, Lu Yuan, and
  Lei Zhang.
\newblock Dynamic head: Unifying object detection heads with attentions.
\newblock In {\em CVPR}, 2021.

\bibitem{dai2021dynamic}
Xiyang Dai, Yinpeng Chen, Jianwei Yang, Pengchuan Zhang, Lu Yuan, and Lei
  Zhang.
\newblock Dynamic detr: End-to-end object detection with dynamic attention.
\newblock In {\em ICCV}, 2021.

\bibitem{damen2018scaling}
Dima Damen, Hazel Doughty, Giovanni~Maria Farinella, Sanja Fidler, Antonino
  Furnari, Evangelos Kazakos, Davide Moltisanti, Jonathan Munro, Toby Perrett,
  Will Price, et~al.
\newblock Scaling egocentric vision: The epic-kitchens dataset.
\newblock In {\em ECCV}, 2018.

\bibitem{dong2021solq}
Bin Dong, Fangao Zeng, Tiancai Wang, Xiangyu Zhang, and Yichen Wei.
\newblock Solq: Segmenting objects by learning queries.
\newblock {\em NeurIPS}, 2021.

\bibitem{dong2021semi}
Guimin Dong, Mingyue Tang, Lihua Cai, Laura~E Barnes, and Mehdi Boukhechba.
\newblock Semi-supervised graph instance transformer for mental health
  inference.
\newblock In {\em ICMLA}, 2021.

\bibitem{dong2022graph}
Guimin Dong, Mingyue Tang, Zhiyuan Wang, Jiechao Gao, Sikun Guo, Lihua Cai,
  Robert Gutierrez, Bradford Campbell, Laura~E Barnes, and Mehdi Boukhechba.
\newblock Graph neural networks in iot: A survey.
\newblock {\em ACM Transactions on Sensor Networks (TOSN)}, 2022.

\bibitem{duan2019centernet}
Kaiwen Duan, Song Bai, Lingxi Xie, Honggang Qi, Qingming Huang, and Qi Tian.
\newblock Centernet: Keypoint triplets for object detection.
\newblock In {\em ICCV}, 2019.

\bibitem{duan2020corner}
Kaiwen Duan, Lingxi Xie, Honggang Qi, Song Bai, Qingming Huang, and Qi Tian.
\newblock Corner proposal network for anchor-free, two-stage object detection.
\newblock In {\em ECCV}, 2020.

\bibitem{Fang_2021_ICCV}
Yuxin Fang, Shusheng Yang, Xinggang Wang, Yu Li, Chen Fang, Ying Shan, Bin
  Feng, and Wenyu Liu.
\newblock Instances as queries.
\newblock In {\em ICCV}, 2021.

\bibitem{Gao_2018_CVPR}
Mingfei Gao, Ruichi Yu, Ang Li, Vlad~I. Morariu, and Larry~S. Davis.
\newblock Dynamic zoom-in network for fast object detection in large images.
\newblock In {\em CVPR}, 2018.

\bibitem{gao2021fast}
Peng Gao, Minghang Zheng, Xiaogang Wang, Jifeng Dai, and Hongsheng Li.
\newblock Fast convergence of detr with spatially modulated co-attention.
\newblock In {\em ICCV}, 2021.

\bibitem{gao2022adamixer}
Ziteng Gao, Limin Wang, Bing Han, and Sheng Guo.
\newblock Adamixer: A fast-converging query-based object detector.
\newblock In {\em CVPR}, 2022.

\bibitem{ghiasi2019fpn}
Golnaz Ghiasi, Tsung-Yi Lin, and Quoc~V Le.
\newblock Nas-fpn: Learning scalable feature pyramid architecture for object
  detection.
\newblock In {\em CVPR}, 2019.

\bibitem{girshick2015fast}
Ross Girshick.
\newblock Fast r-cnn.
\newblock In {\em ICCV}, 2015.

\bibitem{gong2021temporal}
Tao Gong, Kai Chen, Xinjiang Wang, Qi Chu, Feng Zhu, Dahua Lin, Nenghai Yu, and
  Huamin Feng.
\newblock Temporal roi align for video object recognition.
\newblock In {\em AAAI}, 2021.

\bibitem{han2016seq}
Wei Han, Pooya Khorrami, Tom~Le Paine, Prajit Ramachandran, Mohammad
  Babaeizadeh, Honghui Shi, Jianan Li, Shuicheng Yan, and Thomas~S Huang.
\newblock Seq-nms for video object detection.
\newblock {\em arXiv preprint arXiv:1602.08465}, 2016.

\bibitem{he2017mask}
Kaiming He, Georgia Gkioxari, Piotr Doll{\'a}r, and Ross Girshick.
\newblock Mask r-cnn.
\newblock In {\em ICCV}, 2017.

\bibitem{he2016deep}
Kaiming He, Xiangyu Zhang, Shaoqing Ren, and Jian Sun.
\newblock Deep residual learning for image recognition.
\newblock In {\em CVPR}, 2016.

\bibitem{he2021end}
Lu He, Qianyu Zhou, Xiangtai Li, Li Niu, Guangliang Cheng, Xiao Li, Wenxuan
  Liu, Yunhai Tong, Lizhuang Ma, and Liqing Zhang.
\newblock End-to-end video object detection with spatial-temporal transformers.
\newblock In {\em ACM Multimedia}, 2021.

\bibitem{hong2022dynamic}
Qinghang Hong, Fengming Liu, Dong Li, Ji Liu, Lu Tian, and Yi Shan.
\newblock Dynamic sparse r-cnn.
\newblock In {\em CVPR}, 2022.

\bibitem{hu2021istr}
Jie Hu, Liujuan Cao, Yao Lu, ShengChuan Zhang, Yan Wang, Ke Li, Feiyue Huang,
  Ling Shao, and Rongrong Ji.
\newblock Istr: End-to-end instance segmentation with transformers.
\newblock {\em arXiv preprint arXiv:2105.00637}, 2021.

\bibitem{jia2022detrs}
Ding Jia, Yuhui Yuan, Haodi He, Xiaopei Wu, Haojun Yu, Weihong Lin, Lei Sun,
  Chao Zhang, and Han Hu.
\newblock Detrs with hybrid matching.
\newblock {\em arXiv preprint arXiv:2207.13080}, 2022.

\bibitem{jiang2020learning}
Zhengkai Jiang, Yu Liu, Ceyuan Yang, Jihao Liu, Peng Gao, Qian Zhang, Shiming
  Xiang, and Chunhong Pan.
\newblock Learning where to focus for efficient video object detection.
\newblock In {\em ECCV}, 2020.

\bibitem{kang2017t}
Kai Kang, Hongsheng Li, Junjie Yan, Xingyu Zeng, Bin Yang, Tong Xiao, Cong
  Zhang, Zhe Wang, Ruohui Wang, Xiaogang Wang, et~al.
\newblock T-cnn: Tubelets with convolutional neural networks for object
  detection from videos.
\newblock {\em IEEE Transactions on Circuits and Systems for Video Technology},
  28(10):2896--2907, 2017.

\bibitem{Kang_2016}
Kai Kang, Wanli Ouyang, Hongsheng Li, and Xiaogang Wang.
\newblock Object detection from video tubelets with convolutional neural
  networks.
\newblock {\em CVPR}, 2016.

\bibitem{kim2020video}
Dahun Kim, Sanghyun Woo, Joon-Young Lee, and In~So Kweon.
\newblock Video panoptic segmentation.
\newblock In {\em CVPR}, 2020.

\bibitem{law2018cornernet}
Hei Law and Jia Deng.
\newblock Cornernet: Detecting objects as paired keypoints.
\newblock In {\em ECCV}, 2018.

\bibitem{lee2016multi}
Byungjae Lee, Enkhbayar Erdenee, Songguo Jin, Mi~Young Nam, Young~Giu Jung, and
  Phill~Kyu Rhee.
\newblock Multi-class multi-object tracking using changing point detection.
\newblock In {\em ECCV}, 2016.

\bibitem{li2021dynamic}
Changlin Li, Guangrun Wang, Bing Wang, Xiaodan Liang, Zhihui Li, and Xiaojun
  Chang.
\newblock Dynamic slimmable network.
\newblock In {\em CVPR}, 2021.

\bibitem{li2022dn}
Feng Li, Hao Zhang, Shilong Liu, Jian Guo, Lionel~M Ni, and Lei Zhang.
\newblock Dn-detr: Accelerate detr training by introducing query denoising.
\newblock In {\em CVPR}, 2022.

\bibitem{lin2017feature}
Tsung-Yi Lin, Piotr Doll{\'a}r, Ross Girshick, Kaiming He, Bharath Hariharan,
  and Serge Belongie.
\newblock Feature pyramid networks for object detection.
\newblock In {\em CVPR}, 2017.

\bibitem{lin2017focal}
Tsung-Yi Lin, Priya Goyal, Ross Girshick, Kaiming He, and Piotr Doll{\'a}r.
\newblock Focal loss for dense object detection.
\newblock In {\em ICCV}, 2017.

\bibitem{lin2014microsoft}
Tsung-Yi Lin, Michael Maire, Serge Belongie, James Hays, Pietro Perona, Deva
  Ramanan, Piotr Doll{\'a}r, and C~Lawrence Zitnick.
\newblock Microsoft coco: Common objects in context.
\newblock In {\em ECCV}, 2014.

\bibitem{liu2021visual}
Dongfang Liu, Yiming Cui, Xiaolei Guo, Wei Ding, Baijian Yang, and Yingjie
  Chen.
\newblock Visual localization for autonomous driving: Mapping the accurate
  location in the city maze.
\newblock In {\em ICPR}, 2021.

\bibitem{liu2021sg}
Dongfang Liu, Yiming Cui, Wenbo Tan, and Yingjie Chen.
\newblock Sg-net: Spatial granularity network for one-stage video instance
  segmentation.
\newblock In {\em CVPR}, 2021.

\bibitem{liu2019adaptive}
Songtao Liu, Di Huang, and Yunhong Wang.
\newblock Adaptive nms: Refining pedestrian detection in a crowd.
\newblock In {\em CVPR}, 2019.

\bibitem{liu2022dab}
Shilong Liu, Feng Li, Hao Zhang, Xiao Yang, Xianbiao Qi, Hang Su, Jun Zhu, and
  Lei Zhang.
\newblock Dab-detr: Dynamic anchor boxes are better queries for detr.
\newblock {\em arXiv preprint arXiv:2201.12329}, 2022.

\bibitem{liu2016ssd}
Wei Liu, Dragomir Anguelov, Dumitru Erhan, Christian Szegedy, Scott Reed,
  Cheng-Yang Fu, and Alexander~C Berg.
\newblock Ssd: Single shot multibox detector.
\newblock In {\em ECCV}, 2016.

\bibitem{Meng_2021_ICCV}
Depu Meng, Xiaokang Chen, Zejia Fan, Gang Zeng, Houqiang Li, Yuhui Yuan, Lei
  Sun, and Jingdong Wang.
\newblock Conditional detr for fast training convergence.
\newblock In {\em ICCV}, 2021.

\bibitem{ming2021dynamic}
Qi Ming, Zhiqiang Zhou, Lingjuan Miao, Hongwei Zhang, and Linhao Li.
\newblock Dynamic anchor learning for arbitrary-oriented object detection.
\newblock In {\em AAAI}, 2021.

\bibitem{neubeck2006efficient}
Alexander Neubeck and Luc Van~Gool.
\newblock Efficient non-maximum suppression.
\newblock In {\em ICPR}, 2006.

\bibitem{piao2022accloc}
Zhengquan Piao, Junbo Wang, Linbo Tanga, Baojun Zhao, and Wenzheng Wang.
\newblock Accloc: Anchor-free and two-stage detector for accurate object
  localization.
\newblock {\em Pattern Recognition}, 126:108523, 2022.

\bibitem{redmon2016you}
Joseph Redmon, Santosh Divvala, Ross Girshick, and Ali Farhadi.
\newblock You only look once: Unified, real-time object detection.
\newblock In {\em CVPR}, 2016.

\bibitem{redmon2017yolo9000}
Joseph Redmon and Ali Farhadi.
\newblock Yolo9000: better, faster, stronger.
\newblock In {\em CVPR}, 2017.

\bibitem{redmon2018yolov3}
Joseph Redmon and Ali Farhadi.
\newblock Yolov3: An incremental improvement.
\newblock {\em arXiv preprint arXiv:1804.02767}, 2018.

\bibitem{ren2015faster}
Shaoqing Ren, Kaiming He, Ross Girshick, and Jian Sun.
\newblock Faster r-cnn: Towards real-time object detection with region proposal
  networks.
\newblock In {\em NeurIPS}, pages 91--99, 2015.

\bibitem{roh2021sparse}
Byungseok Roh, JaeWoong Shin, Wuhyun Shin, and Saehoon Kim.
\newblock Sparse detr: Efficient end-to-end object detection with learnable
  sparsity.
\newblock {\em arXiv preprint arXiv:2111.14330}, 2021.

\bibitem{russakovsky2015imagenet}
Olga Russakovsky, Jia Deng, Hao Su, Jonathan Krause, Sanjeev Satheesh, Sean Ma,
  Zhiheng Huang, Andrej Karpathy, Aditya Khosla, Michael Bernstein, et~al.
\newblock Imagenet large scale visual recognition challenge.
\newblock {\em International journal of computer vision}, 115(3):211--252,
  2015.

\bibitem{sabater2020robust}
Alberto Sabater, Luis Montesano, and Ana~C Murillo.
\newblock Robust and efficient post-processing for video object detection.
\newblock In {\em IROS}, 2020.

\bibitem{shafiee2017fast}
Mohammad~Javad Shafiee, Brendan Chywl, Francis Li, and Alexander Wong.
\newblock Fast yolo: A fast you only look once system for real-time embedded
  object detection in video.
\newblock {\em arXiv preprint arXiv:1709.05943}, 2017.

\bibitem{song2020fine}
Lin Song, Yanwei Li, Zhengkai Jiang, Zeming Li, Hongbin Sun, Jian Sun, and
  Nanning Zheng.
\newblock Fine-grained dynamic head for object detection.
\newblock {\em NeurIPS}, 2020.

\bibitem{sun2021sparse}
Peize Sun, Rufeng Zhang, Yi Jiang, Tao Kong, Chenfeng Xu, Wei Zhan, Masayoshi
  Tomizuka, Lei Li, Zehuan Yuan, Changhu Wang, et~al.
\newblock Sparse r-cnn: End-to-end object detection with learnable proposals.
\newblock In {\em CVPR}, 2021.

\bibitem{tian2019fcos}
Zhi Tian, Chunhua Shen, Hao Chen, and Tong He.
\newblock {FCOS}: Fully convolutional one-stage object detection.
\newblock In {\em ICCV}, 2019.

\bibitem{tian2021fcos}
Zhi Tian, Chunhua Shen, Hao Chen, and Tong He.
\newblock {FCOS}: A simple and strong anchor-free object detector.
\newblock {\em IEEE Transactions on Pattern Analysis and Machine Intelligence},
  2021.

\bibitem{van2008visualizing}
Laurens Van~der Maaten and Geoffrey Hinton.
\newblock Visualizing data using t-sne.
\newblock {\em Journal of machine learning research}, 9(11), 2008.

\bibitem{voigtlaender2019mots}
Paul Voigtlaender, Michael Krause, Aljosa Osep, Jonathon Luiten, Berin
  Balachandar~Gnana Sekar, Andreas Geiger, and Bastian Leibe.
\newblock Mots: Multi-object tracking and segmentation.
\newblock In {\em CVPR}, 2019.

\bibitem{wang2022ptseformer}
Han Wang, Jun Tang, Xiaodong Liu, Shanyan Guan, Rong Xie, and Li Song.
\newblock Ptseformer: Progressive temporal-spatial enhanced transformer towards
  video object detection.
\newblock {\em arXiv preprint arXiv:2209.02242}, 2022.

\bibitem{wang2020fcos}
Ning Wang, Yang Gao, Hao Chen, Peng Wang, Zhi Tian, Chunhua Shen, and Yanning
  Zhang.
\newblock Nas-fcos: Fast neural architecture search for object detection.
\newblock In {\em CVPR}, 2020.

\bibitem{wang2021fcos}
Ning Wang, Yang Gao, Hao Chen, Peng Wang, Zhi Tian, Chunhua Shen, and Yanning
  Zhang.
\newblock Nas-fcos: Efficient search for object detection architectures.
\newblock {\em International Journal of Computer Vision}, 129(12):3299--3312,
  2021.

\bibitem{wang2018fully}
Shiyao Wang, Yucong Zhou, Junjie Yan, and Zhidong Deng.
\newblock Fully motion-aware network for video object detection.
\newblock In {\em ECCV}, 2018.

\bibitem{wang2019dynamic}
Yue Wang, Yongbin Sun, Ziwei Liu, Sanjay~E Sarma, Michael~M Bronstein, and
  Justin~M Solomon.
\newblock Dynamic graph cnn for learning on point clouds.
\newblock {\em Acm Transactions On Graphics (tog)}, 38(5):1--12, 2019.

\bibitem{wang2022anchor}
Yingming Wang, Xiangyu Zhang, Tong Yang, and Jian Sun.
\newblock Anchor detr: Query design for transformer-based detector.
\newblock In {\em AAAI}, 2022.

\bibitem{wu2019sequence}
Haiping Wu, Yuntao Chen, Naiyan Wang, and Zhaoxiang Zhang.
\newblock Sequence level semantics aggregation for video object detection.
\newblock In {\em ICCV}, 2019.

\bibitem{xu2018youtube}
Ning Xu, Linjie Yang, Yuchen Fan, Dingcheng Yue, Yuchen Liang, Jianchao Yang,
  and Thomas Huang.
\newblock Youtube-vos: A large-scale video object segmentation benchmark.
\newblock {\em arXiv preprint arXiv:1809.03327}, 2018.

\bibitem{yang2019video}
Linjie Yang, Yuchen Fan, and Ning Xu.
\newblock Video instance segmentation.
\newblock In {\em ICCV}, 2019.

\bibitem{yao2020sm}
Lewei Yao, Hang Xu, Wei Zhang, Xiaodan Liang, and Zhenguo Li.
\newblock Sm-nas: Structural-to-modular neural architecture search for object
  detection.
\newblock In {\em AAAI}, 2020.

\bibitem{yao2021efficient}
Zhuyu Yao, Jiangbo Ai, Boxun Li, and Chi Zhang.
\newblock Efficient detr: improving end-to-end object detector with dense
  prior.
\newblock {\em arXiv preprint arXiv:2104.01318}, 2021.

\bibitem{yu2019autoslim}
Jiahui Yu and Thomas Huang.
\newblock Autoslim: Towards one-shot architecture search for channel numbers.
\newblock {\em arXiv preprint arXiv:1903.11728}, 2019.

\bibitem{yu2019universally}
Jiahui Yu and Thomas~S Huang.
\newblock Universally slimmable networks and improved training techniques.
\newblock In {\em ICCV}, 2019.

\bibitem{zhang2020dynamic}
Hongkai Zhang, Hong Chang, Bingpeng Ma, Naiyan Wang, and Xilin Chen.
\newblock Dynamic r-cnn: Towards high quality object detection via dynamic
  training.
\newblock In {\em ECCV}, 2020.

\bibitem{zhang2022dino}
Hao Zhang, Feng Li, Shilong Liu, Lei Zhang, Hang Su, Jun Zhu, Lionel~M Ni, and
  Heung-Yeung Shum.
\newblock Dino: Detr with improved denoising anchor boxes for end-to-end object
  detection.
\newblock {\em arXiv preprint arXiv:2203.03605}, 2022.

\bibitem{zhang2021detr}
Jingyi Zhang, Jiaxing Huang, Zhipeng Luo, Gongjie Zhang, and Shijian Lu.
\newblock Da-detr: Domain adaptive detection transformer by hybrid attention.
\newblock {\em arXiv preprint arXiv:2103.17084}, 2021.

\bibitem{zhang2021dynamic}
Miao Zhang, Jie Liu, Yifei Wang, Yongri Piao, Shunyu Yao, Wei Ji, Jingjing Li,
  Huchuan Lu, and Zhongxuan Luo.
\newblock Dynamic context-sensitive filtering network for video salient object
  detection.
\newblock In {\em ICCV}, 2021.

\bibitem{zhang2021studying}
Zhengming Zhang and Renran Tian.
\newblock Studying battery range and range anxiety for electric vehicles based
  on real travel demands.
\newblock In {\em Proceedings of the human factors and ergonomics society
  annual meeting}, volume~65, pages 332--336. SAGE Publications Sage CA: Los
  Angeles, CA, 2021.

\bibitem{zhang2022attention}
Zhengming Zhang, Renran Tian, Rini Sherony, Joshua Domeyer, and Zhengming Ding.
\newblock Attention-based interrelation modeling for explainable automated
  driving.
\newblock {\em IEEE Transactions on Intelligent Vehicles}, 2022.

\bibitem{zhou2022transvod}
Qianyu Zhou, Xiangtai Li, Lu He, Yibo Yang, Guangliang Cheng, Yunhai Tong,
  Lizhuang Ma, and Dacheng Tao.
\newblock Transvod: End-to-end video object detection with spatial-temporal
  transformers.
\newblock {\em arXiv preprint arXiv:2201.05047}, 2022.

\bibitem{zhou2019bottom}
Xingyi Zhou, Jiacheng Zhuo, and Philipp Krahenbuhl.
\newblock Bottom-up object detection by grouping extreme and center points.
\newblock In {\em CVPR}, 2019.

\bibitem{zhu2020dynamic}
Mingjian Zhu, Kai Han, Changbin Yu, and Yunhe Wang.
\newblock Dynamic feature pyramid networks for object detection.
\newblock {\em arXiv preprint arXiv:2012.00779}, 2020.

\bibitem{zhu2021deformable}
Xizhou Zhu, Weijie Su, Lewei Lu, Bin Li, Xiaogang Wang, and Jifeng Dai.
\newblock Deformable detr: Deformable transformers for end-to-end object
  detection.
\newblock In {\em ICLR}, 2021.

\bibitem{zhu2017flow}
Xizhou Zhu, Yujie Wang, Jifeng Dai, Lu Yuan, and Yichen Wei.
\newblock Flow-guided feature aggregation for video object detection.
\newblock In {\em ICCV}, 2017.

\bibitem{zhu2017deep}
Xizhou Zhu, Yuwen Xiong, Jifeng Dai, Lu Yuan, and Yichen Wei.
\newblock Deep feature flow for video recognition.
\newblock In {\em CVPR}, 2017.

\end{thebibliography}
}

\end{document}